\documentclass{article}
\usepackage{ijcai17}

\usepackage{times}
\usepackage{color}
\usepackage{graphicx} 
\usepackage{subfigure}
\usepackage{algorithm}
\usepackage{algorithmic}
\usepackage{amsmath} 
\usepackage{amssymb} 
\usepackage{amsfonts}
\usepackage{amsthm}
\usepackage{subfigure}
\usepackage[small]{caption}
\usepackage{url}
\newtheorem{theorem}{Theorem}
\newtheorem{lemma}{Lemma}

\DeclareMathOperator{\tr}{tr}

\DeclareMathOperator{\sgn}{sgn}

\renewcommand{\vec}[1]{\mathbf{#1}}

\title{Theoretic Analysis and Extremely Easy Algorithms \\for Domain Adaptive Feature Learning
}

\author{Wenhao Jiang$^{1}$\thanks{Corresponding author. Email: cswhjiang@gmail.com}, Cheng Deng$^{2}$, Wei Liu$^{1}$, Feiping Nie$^{3}$,  Fu-lai Chung$^{4}$, Heng Huang$^{5,2}$ \\
$^{1}$Tencent AI Lab, P. R. China\\
 $^{2}$Xidian University, P. R. China\\
  $^{3}$Northwestern Polytechnical University, P. R. China \\
 $^{4}$Hong Kong Polytechnic University, Hong Kong\\
  $^{5}$University of Texas at Arlington, USA
}
\begin{document}

\maketitle

\begin{abstract}
Domain adaptation problems arise in a variety of applications, where a training dataset from the \textit{source} domain and a test dataset from the \textit{target} domain typically follow different distributions. The primary difficulty in designing effective learning models to solve such problems lies in how to bridge the gap between the source and target distributions. In this paper, we provide comprehensive analysis of feature learning algorithms used in conjunction with linear classifiers for domain adaptation. Our analysis shows that in order to achieve good adaptation performance, the second moments of the source domain distribution and target domain distribution should be similar. Based on our new analysis, a novel extremely easy feature learning algorithm for domain adaptation is proposed. Furthermore, our algorithm is extended by leveraging multiple layers, leading to a deep linear model. We evaluate the effectiveness of the proposed algorithms in terms of domain adaptation tasks on the Amazon review dataset and the spam dataset from the ECML/PKDD 2006 discovery challenge.
\end{abstract}

\section{Introduction}
In traditional supervised learning, it is assumed that training data samples and test data samples are drawn from the same distribution when a learner (e.g., classifier) is trained. However, this may not be valid in domain adaptation problems. We often have plenty of labeled examples from one domain (\textit{source} domain) for training a classifier and intend to apply the trained classifier to another different domain (\textit{target} domain) with very few or even no labeled example. Domain adaptation has tight relationship with semi-supervised learning \cite{liu2012robust} and has been studied before under different names, including covariate shift~\cite{Shimodaira2000} and sample selection bias~\cite{Heckman1979,Zadrozny2004}.


The difficulty of domain adaptation comes from the gap between the source distribution and target distribution. Hence, the learning model trained in the source domain could not be directly used in the target domain. Many algorithms have been proposed to address domain adaptation problems. For example, a semi-supervised method, namely structural correspondence learning (SCL), was proposed in~\cite{Blitzer2006}. SCL defines pivot features which are common to both domains and tries to find the correlation between pivot and non-pivot features. It extracts the corresponding subspace and augments the original feature space with the subspace towards more effective classification. The method in~\cite{Blitzer2011} tries to find the correlations between original features and identify the subspaces in which a good predictor could be obtained by training a classifier merely on the samples from the source domain. The kernel mean matching (KMM) method was proposed in~\cite{Huang2007}, which  aims to minimize the distance between means of the training and test samples in a reproducing kernel Hilbert space (RKHS) by reweighting the samples from the source domain.  In~\cite{Chen2011g}, a co-training method called CODA for domain adaptation was proposed. In each iteration, CODA formulates an individual optimization problem which simultaneously learns a target predictor, a split of the feature space into views, and a subset of source and target features to be included in the predictor. It tries to progressively bridge the gap between the source and target domains by adding both the features and instances that the current predictor is most confident about. 
Nonnegative matrix factorization was employed to bridge the gap between domains by sharing feature clusters \cite{wang2011dyadic}. 
Glorot \emph{et al.}~\cite{Glorot2011} proposed to learn robust feature representations with stacked denoising auto-encoders (SDA) for domain adaptation. Marginalized stacked denoising auto-encoder (mSDA)~\cite{Chen2012a},  a variant of SDA with a slightly different network structure, was proposed to address the drawback of SDA being too slow in training. Chen \emph{et al.}~\cite{Chen2012a} noticed that the random feature corruption for SDA can be marginalized out, which is equivalent to training a learning model with an infinitely large number of corrupted input data conceptually. Moreover, the linear denoising auto-encoders used in SDA have a closed form, which help speed up the computation. Promising performance was achieved in cross-domain sentiment analysis tasks. In \cite{Sun2015}, correlation alignment (CORAL) exploits the difference between the covariance matrices as a measure of the gap between the source and target domains. A shallow linear feature algorithm was then proposed. Inspired by the theory of domain adaptation \cite{Ben-David2007}, domain-adversarial neural networks (DANN) \cite{Ajakan2014} and domain adaptation by back-propagation (DAB) \cite{Ganin2015} optimize an approximated domain distance and an empirical training error on the source domain to seek the hidden representations for both source and target domain samples. DANN and DAB can be treated as classifiers for tackling domain adaptation problems, and also be trained on top of other feature learning algorithms, \emph{e.g.,} mSDA.

Besides the aforementioned methods, domain adaptation has also been studied theoretically. Most of them are built on distances measuring the dissimilarity between different distributions, and generalization bounds are derived based on the proposed distances. In~\cite{Ben-David2007}, the $\mathcal{A}$-distance was used to analyze representations for domain adaptation, and VC-dimension shaped generalization bounds were derived for domain adaptation. The analysis showed that a good feature learning algorithm should achieve a low training error on source domain and a small $\mathcal{A}$-distance simultaneously. In \cite{Blitzer2008}, a uniform convergence bound was provided and it was extended to domain adaptation with multiple source domains combined with weights. In~\cite{Mansour2009,Zhang2013}, the $\mathcal{A}$-distance was extended to a more general form and could be used to compare the distance for more general tasks such as regression. All the distances mentioned above were defined in a worst-case sense. In \cite{Germain2013}, a distance defined in an average sense by making use of PAC-Bayesian theory was suggested and an algorithm that simultaneously optimizes the error on the source domain, the hypothesis complexity, and the distance was proposed. In~\cite{Ben-David2010a}, the conditions for the success of domain adaptation were analyzed. It was shown that a small distance between source and target domains and the existence of a low error classifier on both domains in the hypothesis class are necessary for the success of domain adaptation.

In this paper, we provide theoretic analysis of several key issues pertaining to effective domain adaptive feature learning, showing that the difference (measured by Frobenius norm) between the second moments of source and target domain distributions should be small. Based on this analysis, we propose a simple yet effective feature learning algorithm used in conjunction with linear classifiers. To further improve the feature learning quality, we employ a deep learning approach inspired by stacked denoising auto-encoders in~\cite{Glorot2011,Chen2012a}, leading to a deep linear model (DLM). DLMs are easy to analyze and usually are the starting point for theoretical development of neural networks \cite{Goodfellow-et-al-2016}. 
 Finally, we demonstrate the effectiveness of our proposed feature learning algorithms on the Amazon review and spam datasets.

\vspace{-2mm}
\section{Theoretic Analysis and New Algorithms}
\subsection{Notations and Background}
Usually,  a domain is considered as a pair consisting of a distribution $\mathcal{D}$ on $\mathcal{X}$ and a labeling  function $f:  \mathcal{X} \mapsto \{0, 1\}$. In this paper, we consider two domains, a \textit{source} domain $\langle \mathcal{D}_S, f_S\rangle$ and a \textit{target} domain $\langle \mathcal{D}_T, f_T\rangle$.  The probability density functions for source  distribution and target distribution are $p_S(\vec{x})$ and $p_{T}(\vec{x})$, respectively. There exist $n_s$ samples $D_{S} = \{\mathbf{x}_1,\cdots, \mathbf{x}_{n_s}\} \in \mathbb{ R}^{d}$  with labels $L_{S} = \{ y_1,\cdots, y_{n_s}\}$  sampled from the source domain while  $n_t$ samples $D_T = \{\mathbf{x}_{n_s+1}, \cdots, \mathbf{x}_n  \}\in \mathbb{ R}^{d}$ from the \textit{target} domain  are sampled without labels. They have the same number of features, which is denoted as $d$. The samples from source domain form the data matrix $X_S \in \mathbb{R}^{d \times n_s }$, and samples from target domain form $X_T \in \mathbb{R}^{d \times n_t }$. We use  $X \in \mathbb{ R}^{d\times n}$ to denote the data matrix containing the samples from both domains.

A hypothesis is a function $h :\mathcal{X} \mapsto \{0, 1\}$. The risk of a hypothesis $h$ over domain $\langle \mathcal{D}, f\rangle$  is denoted as $ \epsilon(h,f) = E  [|h(\vec{x}) - f(\vec{x})|]$, which is the difference between the two functions. We use  notations
 $\epsilon_{S}(h,f_S) = E_{S} [|h(\vec{x}) - f_S(\vec{x})|] $ and 
 $\epsilon_{T}(h,f_T) = E_{T} [|h(\vec{x}) - f_T(\vec{x})|] $
to denote the risk of $h$ on source domain and target domain  respectively. In this paper, we consider only linear classifiers.  We denote the best linear classifier for source domain as $h^*_{S}$ with corresponding parameter vector $\vec{w}^*_{S}$, the best classifier for target domain as $h^*_T$ with parameter vector $\vec{w}_{T}^*$. For simplicity of expression, we define the difference of expectation of a function $t(\vec{x})$ on source domain and target domain as $E_{S-T}[t(\vec{x})] = E_{S}[t(\vec{x})] - E_{T}[t(\vec{x})]
 =\int t(\vec{x})(p_S(\vec{x}) - p_T(\vec{x})) d \vec{x}.$

Our goal in this paper is to learn a new feature representation with samples from \textit{source} domain and \textit{target} domain such that we can train a classifier  with the learned representations on source domain and apply it directly to target domain  for achieving low  $\epsilon_{T}(h,f_T)$ value.
\vspace{-1mm}
\subsection{Analysis of Feature Learning for Linear Classifiers}

\paragraph{Assumption.} We assume that $\|\vec{x}\|\leq \alpha $ for all samples and  there exists a low-risk linear classifier $h^*$  with parameter $\vec{w}^*$ for both domains. Its risk on source domain is $\lambda_1 = \epsilon_S(h^*, f_S)$, and risk on target domain is $\lambda_2 = \epsilon_T(h^*, f_T)$ and $\lambda = \epsilon_S(h^*, f_S) + \epsilon_T(h^*, f_T)$, where $\lambda$ is assumed to be small.


With triangle inequality, it was shown that for any classifier $h$, the following is satisfied~\cite{Blitzer2008}:
\begin{align*}
\epsilon_T(h,f_T)  & \leq  \  \epsilon_{T}(h^*,f_T)  +  \epsilon_{S}(h^*,f_S) + \epsilon_{S}(h,f_S)+ \\
& \quad \quad |\epsilon_{S}(h,h^*) - \epsilon_{T}(h,h^*)| \nonumber \\
&= \lambda + \epsilon_{S}(h,f_S)+ |\epsilon_{S}(h,h^*) - \epsilon_{T}(h,h^*)| \nonumber \\
&= \lambda + \epsilon_{S}(h,f_S) + \\
& \quad \quad |E_{S} [|h(\vec{x}) - h^*(\vec{x})|] -  E_{T} [|h(\vec{x}) - h^*(\vec{x})|] | \nonumber
  \end{align*}
From this inequality, we can see that the performance of classifier $h$ on target domain is determined by: 1) the risk of the best classifier $h^*$ on both domains; 2) the risk of $h$ on source  domain; 3) the difference of dissimilarity between $h$ and $h^*$ on both domains. A good feature learning algorithm should decrease the sum of these three terms. In this paper, we focus on the third part and provide an analysis for effective linear feature learning algorithm for linear classifiers.

In the above inequality, $|h(\vec{x}) - h^*(\vec{x})|$  measures the difference between $h(\vec{x})$ and $h^*(\vec{x})$, which is a non-smooth 0-1 loss and is difficult to analyze. In this paper, we use a smooth approximation for the measure of dissimilarity between $h$ and $h^*$. We denote the smooth loss as $l_{\vec{x},y}(z) = l_{\vec{x},y}(\vec{w}^T\vec{x})$, where $\vec{w}$ is the parameter for the linear classifier $h(\vec{x})$ and $y$ is the true label of sample $\vec{x}$. $l_{\vec{x},y}(z)$ can be logistic loss or smooth  approximation of hinge loss \cite{Zhang2001}. We denote $l'_{\vec{x},y}(z) $ and $l''_{\vec{x},y}(z) $ as the first and second derivatives of $l_{\vec{x},y}(z)$ respectively.  Thus, we have the following theorem, which is proved in the supplementary material.

\begin{theorem}
Let $R = \{ \vec{x}=(x_1, \cdots, x_d)^T : a_i < x_i < b_i \ (i = 1,2,\cdots, d) \}$ and $R' = \{ \vec{y}=(\vec{x}^T, \vec{z}^T)^T = (x_1, \cdots, x_d, z_1, \cdots, z_d)^T : a_i < x_i < b_i \ and \ a_i < z_i < b_i \ (i = 1,2,\cdots, d) \}$. Assume that any sample in our problem satisfies  $\vec{x} \in R$, we have the following inequality:
\begin{align}\label{ineq:theorem}
 & \left|E_{S} [ l_{\vec{x},y^*}(\vec{w}^T\vec{x}) ] - E_{T} [ l_{\vec{x},y^*}(\vec{w}^T\vec{x}) ] \right|\nonumber \\
\leq & \ \left|E_{S-T} \left[  l_{\vec{x},y^*}(\vec{x}^T\vec{w}^*) \right]  \right| + \\
   &\quad \frac{1}{2}\|\vec{w} - \vec{w}^*\|^2 \sqrt{CM +\frac{ G^2 }{M}   \left\| E_{S-T}[  \vec{x}  \vec{x}^T ] \right\|_F^2 } \nonumber
\end{align}
 where $ C$, $M$ and $G$ are defined as
 \begin{align*}
C & = \frac{1}{2M^2} \int_{R'}  \sum_{i=1}^{2d} \left(|v(\vec{y})| \|D_i u(\vec{y})\|_{\infty} + \right. \\
& \quad \quad \left. |u(\vec{y})| \|D_i v(\vec{y})\|_{\infty }\right)E_i(\vec{y}) d\vec{y} ,\\	
 M &= \Pi_{i=1}^{d} (b_i - a_i)^2, \ E_i(\vec{y}) = \int_{R'}|y_i - r_i|d \vec{y}, \\
  u(\vec{y}) & = l''_{\vec{x},y^*}(\vec{x}^T\vec{w}^*)
l''_{\vec{x},y^*}(\vec{z}^T\vec{w}^*), \\
v(\vec{y}) &=  \tr \left(   (p_{S}(\vec{x}) - p_{T}(\vec{x}) ) \vec{x}  \vec{x}^T
  (p_{S}(\vec{z}) - p_{T}(\vec{z})) \vec{z}  \vec{z}^T  \right) ,\\
    d\vec{x} &= dx_i\cdots dx_d, \ d\vec{z} = dz_i\cdots dz_d, \\
    G &= \int  l_{\vec{x},y^*}''(\vec{x}^T\vec{w}^*)  d\vec{x},
 \end{align*}
$D_i u(\vec{x})$ and $D_i v(\vec{x})$ are partial derivatives, $y^*$ is the label of sample $\vec{x}$ predicted by $\vec{w}^*$ as $y^* = \sgn(\vec{x}^T\vec{w}^*)$.
\end{theorem}

Our goal is to find new representations such that the  approximation for the measure of dissimilarity between $h$ and $h^*$ is small. Here, we minimize its upper bound instead. In the right hand side of inequality~\eqref{ineq:theorem}, $\vec{w}^*$ is the unknown best possible classifier, hence constants like $C$, $G$ and $\left|E_{S-T} \left[  l_{\vec{x},y^*}(\vec{x}^T\vec{w}^*) \right]  \right|$ which  are related to $\vec{w}^*$ could not be optimized directly by feature learning algorithms. We assume that the learned features lie in the same region with the original feature, hence $M$ could not be optimized. Therefore, in order to achieve good performance, we should minimize $ \|E_{S-T}[  \vec{x}  \vec{x}^T ]\|_{F}^2$. We will use  $ \|E_{S-T}[  \vec{x}  \vec{x}^T ]\|_{F}^2$ as a measure of domain distance and design a feature learning algorithm based on it.

\subsection{Algorithm}
Let us  denote $M_S = \frac{1}{n_s} X_SX_S^T$, $M_T = \frac{1}{n_t} X_TX_T^T$ and  $\Delta M =M_S-M_T $, and our goal is to learn a linear transformation $P$ such that the learned data matrix $P^TX$ is  suitable for domain adaptation. Based on the above analysis, one of the objectives is to  minimize $ \|P^T \Delta MP\|_F^2  =\tr(\Delta M PP^T \Delta MPP^T)$,
which is difficult to optimize. Because positive-semidefinite matrices  $A$ and $B$ satisfy
$0 \leq \tr(AB) \leq \tr(A)\tr(B)$, we have $\tr(\Delta M PP^T\Delta MPP^T) \leq \tr(\Delta M PP^T\Delta M) \tr(PP^T)$.
Therefore, the domain distance is bounded by $\tr(P^T\Delta M^2 P) \tr(PP^T)$. Hence, our goal can be expressed as finding a matrix $P$ with not only small Frobenius norm but also small $\tr(P^T\Delta M^2 P)$. Moreover, the learned representation should also be similar to the original ones. Hence, we can simply use 
\begin{equation}
\min_{P} \|X^T - X^TP\|_F^2 + \gamma_1 \tr(PP^T) + \gamma_2 \tr(P^T\Delta M^2P)	\nonumber
\end{equation}
as objective function\footnote{To express clearly, we omit the bias term.}. In order to force each feature to contribute equally, the length of features should be incorporated by the second term.  The final objective function of our method is expressed as follows
\begin{eqnarray}
\label{eq:obj}
\small
 \min_{P} \|X^T - X^TP\|_F^2 + \gamma_1 \tr(P^T \Lambda P) + \gamma_2 \tr(P^T\Delta M^2P),
\end{eqnarray}
where $\Lambda$ is a diagonal matrix with  $\Lambda_{ii} = X_{i\cdot}X_{i\cdot}^T$, and $X_{i\cdot}$ is the $i$th row of data matrix $X$.   We term our method as  \textbf{F}eature \textbf{L}e\textbf{A}rning with second \textbf{M}oment \textbf{M}atching (FLAMM). We will prove that our method will decrease the distance between domains under some conditions. If $\gamma_2=0$, this algorithm becomes regularized linear regression, which is called simple feature learning algorithm, referred as SFL\footnote{The linear transformation step of mSDA could be rewritten in a form similar to SFL approximately. Hence, each layer of mSDA could be seen as SFL plus a non-linear transformation step.}. SFL does not consider minimizing the distance between the second moments \textit{explicitly}, but it still can improve the adaptation performance in many cases which can be seen from the experimental results section. With our theoretical analysis of domain adaptation, we will  make an attempt to illustrate why this simple method works.

To further improve the performance, we adopt the strategy from deep learning methods and apply this algorithm to the learned feature repeatedly. Following the tradition of deep learning, we call the process of finding one linear transformation matrix and updating the data matrix as one layer. The whole algorithm is summarized in Algorithm~\ref{alg:mom}. We can see that the output of our method has the same dimensionality as input. Moreover, the final output is just the linear transformation of the original data matrix, which makes it easier to analyze than the other auto-encoder based methods.

\vspace{-2mm}
\begin{algorithm}[H]
   \caption{Feature learning with second moment matching (FLAMM).}
   \label{alg:mom}
\begin{algorithmic}
   \STATE {\bfseries Input:} $X$, $\gamma_1 $, $\gamma_2$, number of layers $ K$.
\medskip

\FOR{ $k=1$ {\bfseries to} K }
\STATE{
Compute $P$ by solving problem~\eqref{eq:obj}.

Update $X$ by $X = P^TX$.
} \ENDFOR

\medskip

   \STATE {\bfseries Output:} $X$
\end{algorithmic}
\end{algorithm}
\vspace{-3mm}
In \cite{Sun2015}, a linear feature algorithm called correlation alignment (CORAL) was proposed. It uses the difference between the covariance matrices of source distribution and target distribution as a measure of the gap between source domain and target domain. CORAL just whitens source domain and then recolors it with covariance matrix of target domain. The authors did not provide theoretical analysis for the gap measure used and CORAL is also not a deep feature learning method. We will show that our method is better than CORAL.

\vspace{-1mm}
\subsection{Analysis of Our Method}
In this subsection, we will provide an analysis of FLAMM. Since FLAMM is just a revision of SFL by adding a term, we will analyze SFL first. 
We will prove that the domain distance decreases for each layer of SFL under some conditions, which also explains why mSDA works since the layer of mSDA is similar to that of SFL. Based on that, we will provide an analysis of FLAMM. Here are some necessary lemmas.

\begin{lemma}[Weyl's inequality \cite{Horn2012} Theorem 4.3.1] \label{weyl_inequality}
	Let $A, B \in R^{n \times n}$ be symmetric matrices, and denote the eigenvalue of matrix $A$ as $\lambda_i(A)$, which is arranged in increasing order $\lambda_1(A) \leq \lambda_2(A) \leq \cdots \leq \lambda_{n}(A)$. For $k = 1,2, \cdots, n$, we have
	$
	\lambda_{k}(A) + \lambda_1(B) \leq \lambda_k(A+B) \leq \lambda_k(A) + \lambda_n(B)	
	$.
\end{lemma}

\begin{lemma}[\cite{Horn1991} Theorem 3.3.16] \label{singular_value_estimate}
	Let $A, B $ be $n \times n$ matrices, and denote the singular values of matrix $A$ as $\sigma_i(A)$, which is arranged in increasing order $\sigma_1(A) \leq \sigma_2(A) \leq \cdots \leq \sigma_{n}(A)$. For $k = 1,2, \cdots, n$, we have $\sigma_k(AB) \leq \sigma_k(A) \sigma_n(B)$.	
\end{lemma}

\begin{lemma}\label{psd_cond}
	$P$ is computed by solving problem~\eqref{eq:obj} with $\gamma_2 = 0$, if $\gamma_1 \geq \frac{\lambda_d(XX^T) - \lambda_1(XX^T)}{\lambda_1(\Lambda)}$ then $I-PP^T$ is positive definite matrix.
\end{lemma}
\textbf{Proof:}
According to Lemma~\ref{weyl_inequality}, we have $\lambda_i(XX^T + \gamma_1 \Lambda) \geq \gamma_1 \lambda_i(\Lambda) +\lambda_1(XX^T)$,
therefore $\lambda_{i}((XX^T + \gamma_1 \Lambda)^{-1}) \leq \frac{1}{\gamma_1 \lambda_i(\Lambda) + \lambda_1(XX^T)}$.
By Lemma~\ref{singular_value_estimate}, we have:
\begin{align*}
	 \sigma_{i}(P) =&  \sigma_{i}((XX^T + \gamma_1 \Lambda)^{-1}XX^T)  \\
\leq & \sigma_{i}((XX^T + \gamma_1 \Lambda)^{-1}) \sigma_{d}(XX^T) \\
= &\lambda_{i}((XX^T + \gamma_1 \Lambda)^{-1}) \lambda_{d}(XX^T)  \\
\leq & \frac{\lambda_{d}(XX^T)}{\gamma_1 \lambda_i(\Lambda) + \lambda_1(XX^T)}.
\end{align*}
If $\gamma_1 > \frac{\lambda_d(XX^T) - \lambda_1(XX^T)}{\lambda_1(\Lambda)}$, we will have $\frac{\lambda_{d}(XX^T)}{\gamma_1 \lambda_i(\Lambda) + \lambda_1(XX^T)}  < 1$, which means $\sigma_i(P) < 1$. Hence, $\lambda_i(PP^T) = \sigma_{i}(P)^2 <1$. Therefore,  $I - PP^T$ is positive definite matrix.
\hfill $\Box$

\begin{lemma}
	$A$, $B$, $C$ are symmetric matrices and $ A \succeq 0 $, $B \preceq C$, then $\tr(AB) \leq \tr(AC)$.
\end{lemma}
\textbf{Proof:}
If $A$ is a rank-one matrix, we can express $A$ as $A = vv^T$. Then we have $\tr(AB) = \tr(vv^TB) = v^TBv \leq  v^TCv  = \tr(AC)$. If $A$ is not a rank-one matrix, it can be expressed as $A = \sum_{i} \lambda_i u_i u_i^T$. Since $A \succeq 0$, we denote $v_i = \sqrt{\lambda_i}u_i$ and write $A = \sum_i v_iv_i^T$. Then we have
\begin{eqnarray}
	&\tr(AB)= \tr(\sum_i v_iv_i^T B) =  \sum_i \tr( v_iv_i^T B) \\
	&\quad\quad  \quad\quad  \leq   \sum_i \tr( v_iv_i^T C) =  \tr( \sum_i v_iv_i^T C) =\tr(AC).\nonumber
\end{eqnarray}
\hfill $\Box$

Based on the above lemmas, we have the following theorem:
\begin{theorem}\label{theorem:semi-works}
 If  $P$ is computed by solving problem~\eqref{eq:obj} with $\gamma_2 = 0$ and $\gamma_1 \geq \frac{\lambda_d(XX^T) - \lambda_1(XX^T)}{\lambda_1(\Lambda)}$, then the inequality $\left\| \Delta M \right\|_{F}^2 \geq  \left\|P^T \Delta M  P\right\|_{F}^2$ is satisfied.
\end{theorem}
\textbf{Proof:}
In this proof, we will use the following notations:
\begin{align*}
	M =& XX^T, & M_n = & \Lambda^{-\frac{1}{2}}M\Lambda^{-\frac{1}{2}},\\
	\Delta M_n =& \Lambda^{-\frac{1}{2}}\Delta M\Lambda^{-\frac{1}{2}},  &
	P_n  =& \Lambda^{\frac{1}{2}} P \Lambda^{-\frac{1}{2}} = (M_n + \gamma_1 I)^{-1} M_n.\nonumber
\end{align*}
By Lemma~\ref{psd_cond}, we know that if $\gamma_1 \geq \frac{\lambda_d(XX^T) - \lambda_1(XX^T)}{\lambda_1(\Lambda)}$ we will obtain that $I-PP^T$ is a positive definite matrix, which means $ \Lambda - P_n\Lambda P_n^T $ is also a positive definite matrix. Hence, we have $P_n\Lambda P_n^T \preceq  \Lambda $. Therefore, we obtain:
\begin{align*}
\left\|P^T \Delta M  P\right\|_{F}^2
&= \left\|\Lambda^{\frac{1}{2}} P_n^T \Delta M_n  P_n \Lambda^{\frac{1}{2}}\right\|_{F}^2 \\
&= \tr\left( (\Delta M_n  P_n \Lambda P_n^T \Delta M_n)(P_n \Lambda P_n^T)\right) \\
&\leq  \tr( \Delta M_n  P_n \Lambda P_n^T \Delta M_n \Lambda ) \\
&= \tr\left(   (P_n \Lambda P_n^T )(\Delta M_n \Lambda \Delta M_n)\right) \\
&\leq  \tr(   \Lambda \Delta M_n \Lambda \Delta M_n) \\
&= \|\Delta M\|_{F}^2. 
\end{align*}
\hfill $\Box$
\vspace{-1mm}

From the above theorem, we can see that on a certain layer of SFL, if $\gamma_1$ is big enough distance between the second moments will definitely get smaller compared to that of the input matrix. Hence, SFL will decrease the distance implicitly on one layer. This also explains why mSDA works for domain adaptation problems, since the layer of mSDA is very similar to that of SFL. Therefore, even if SFL and mSDA do not optimize the difference between the two second moment explicitly, they are capable of improving the domain adaptation performance. For FLAMM, with the third term in the objective function, it can decrease the domain distance explicitly.  Hence, FLAMM could be seen as a \textit{trade-off between reconstruction error and domain distance}. And we can achieve the same reconstruction error with smaller domain distance. That's the reason why FLAMM performs better than SFL, which can be seen in the following section. In practice, we do not need to compute the exact condition in theorem~\ref{theorem:semi-works}, we can just treat $\gamma_1$ and $\gamma_2$ as ordinary parameters and select them using the validation set. We will illustrate the changes of distance empirically in the experimental results section.

 \begin{table*}[t] 
\caption{Domain adaptation performance (accuracy $\%$) on the Amazon review dataset. We use the initial letters B, D, E, and K to represent the four domains. The last line is about the average performance over the 12 tasks. }
\vspace{-3mm}
\label{acc_on_sentiment}
\begin{center}
\small
\begin{tabular}{l||c|c|c|c|c|c|c|c|c}
\hline
	&$\textrm{FLAMM}_{2}$  &$\textrm{SFL}_{2}$& $\textrm{FLAMM}_{1}$ & $\textrm{SFL}_{1}$ &   $\textrm{mSDA}_5$& \textit{tf-idf} & PCA& CODA &CORAL\\
\hline  \hline
B $\rightarrow$ D &84.06  &\textbf{84.15}  &83.31 &82.27 &83.96 &79.00 &81.65 & 81.01& 81.46\\
B $\rightarrow$ E &\textbf{81.07}  &77.26  &79.63 &74.97 &77.17 &73.32 &74.83 & 80.83&74.69 \\
B $\rightarrow$ K &83.65  &83.01  &82.86 &82.77 &\textbf{84.11} &76.69 & 81.51&82.63 &76.73 \\
\hline
D $\rightarrow$ B &83.46  &\textbf{83.58}  &82.14 &82.79 &83.35 &77.12 & 80.5& 79.37&79.57 \\
D $\rightarrow$ E & 82.69  &82.01  &80.96 &79.75 &\textbf{82.88} &73.73 &79.65 &81.68 &76.45 \\
D $\rightarrow$ K &\textbf{88.10}  &88.03  &86.85 &86.96 &87.49 &77.94 &81.56 &84.24 &79.37 \\
\hline
E $\rightarrow$ B &\textbf{80.33}  &79.62  &79.31 &78.16 &80.20 &71.34 &78.48 &77.53 &73.36\\
E $\rightarrow$ D &\textbf{80.56}  &80.27  &79.94 &80.04 &79.71 &71.77 &77.77 &77.56 &74.36\\
E $\rightarrow$ K &\textbf{89.16}  &88.56  &88.44 &88.14 &88.43 &84.02 &86.72 &85.89 &86.18\\
\hline
K $\rightarrow$ B &79.02  &\textbf{79.90}  &78.21 &76.69 &77.63 &71.77 &74.75 &76.49 &72.66 \\
K $\rightarrow$ D &\textbf{80.82}  & 80.53 &79.58 &78.03 &78.22 &73.00 &78.02 &78.61 &74.82\\
K $\rightarrow$ E &86.57  &86.67  &86.77 &86.88 &\textbf{86.91} &83.65 &85.35 &85.59 &85.75\\
\hline
AVG               &\textbf{83.29}  &82.79  &82.34 &81.46 &82.51 &76.11 &80.06 &80.95 &77.95\\
\hline
\end{tabular}
\end{center}
\end{table*}

\begin{table*}[t]
\vspace{-3mm}
\caption{Domain adaptation performance (accuracy $\%$) on the spam dataset.}
\vspace{-4mm}
\label{acc_on_spam}
\begin{center}
\small
\begin{tabular}{l||c|c|c|c|c|c|c|c|c}
\hline
	&$\textrm{FLAMM}_{5}$  &$\textrm{SFL}_{5}$& $\textrm{FLAMM}_{1}$ & $\textrm{SFL}_{1}$ &   $\textrm{mSDA}_3$& \textit{tf-idf} & PCA& CODA&CORAL\\
 \hline \hline
 Public $\rightarrow$ U0  &\textbf{91.99}&87.09 &79.79 &65.48 &90.74 &61.73 &66.58  & 90.45&77.58\\
Public $\rightarrow$ U1   &\textbf{94.19}&93.94 &77.88 &73.12 &92.64&70.67 &73.47  &92.95 &80.98\\
Public $\rightarrow$ U2   &95.15&86.90 &85.20 &76.75 &93.8 &74.15 &75.65  &\textbf{95.33} &87\\
 \hline
AVG                      &\textbf{93.78}&89.31 &80.96 &71.78 &92.39  &68.85 & 71.9 &92.91 &81.85\\
\hline
\end{tabular}
\end{center}
\vspace{-5mm}
\end{table*}

\vspace{-2mm}
\section{Experimental Results}
\vspace{-1mm}

\subsection{Datasets}

We evaluate and analyze the proposed methods on the Amazon review dataset \footnote{\url{http://www.cs.jhu.edu/~mdredze/datasets/sentiment/}}~\cite{Blitzer2007},  and the ECML/PKDD 2006 spam dataset \footnote{\url{http://www.ecmlpkdd2006.org/challenge.html}}  \cite{Bickel2008b}. As in~\cite{Blitzer2007}, a smaller subset of the Amazon review dataset which contains reviews of four types of products: \textit{Books}, \textit{DVDs}, \textit{Electronics}, and \textit{Kitchen appliances}, is used. 
In this dataset, each domain consists of 2000 labeled inputs and approximately 4000 unlabeled ones.  We only consider binary classification problem, i.e. whether a review is positive (higher than 3 stars) or negative (3 stars or lower) as in~\cite{Glorot2011,Chen2012a} and use 5000 most frequent terms as features.

The second dataset is from the ECML/PKDD 2006 discovery challenge which is about personalized spam filtering and generalization across related learning tasks.  It contains two tasks: task A and task~B. We adopt the dataset of task A for our comparisons and analysis because it contains more samples. In this dataset, 4000 labeled training samples were collected from publicly available sources, with half of them being \textit{spam} and  the other half being \textit{non-spam}. The testing samples were collected from 3 different user inboxes, say U0, U1 and U2, each of which consists of 2500 samples. Hence, the distributions of  \textit{source} domain and \textit{target} domain are different since they are from different  sources. In this dataset, there are  three adaptation tasks in total.  As in the Amazon review dataset,  5000  most frequent terms were chosen as features. Three samples were deleted as a result of not containing any of these 5000 terms. Hence, we have 7497 testing samples totally.

\begin{figure*}[!t]
  \centering
    \subfigure[Amazon review dataset (B $\rightarrow$ D)]{
    \includegraphics[scale=0.25]{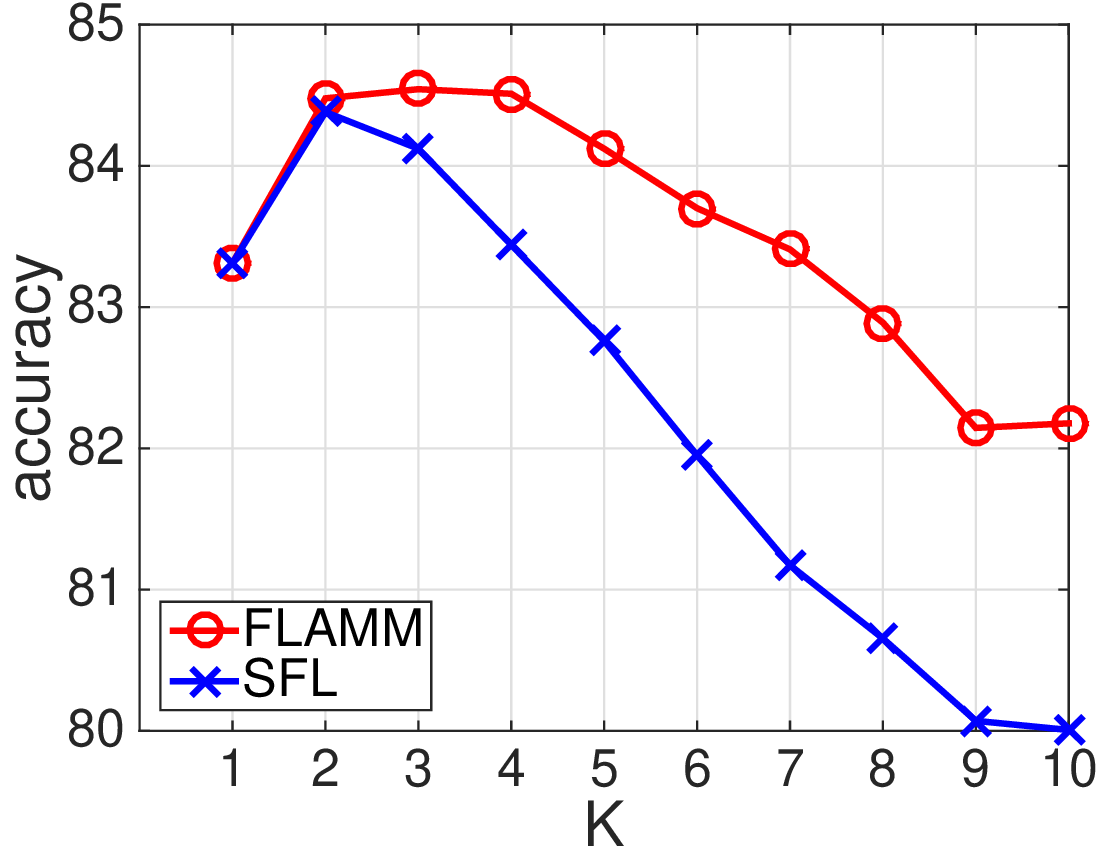}
    }\qquad\qquad
    \subfigure[Spam dataset (Public $\rightarrow$ U0)]{
    \includegraphics[scale=0.25]{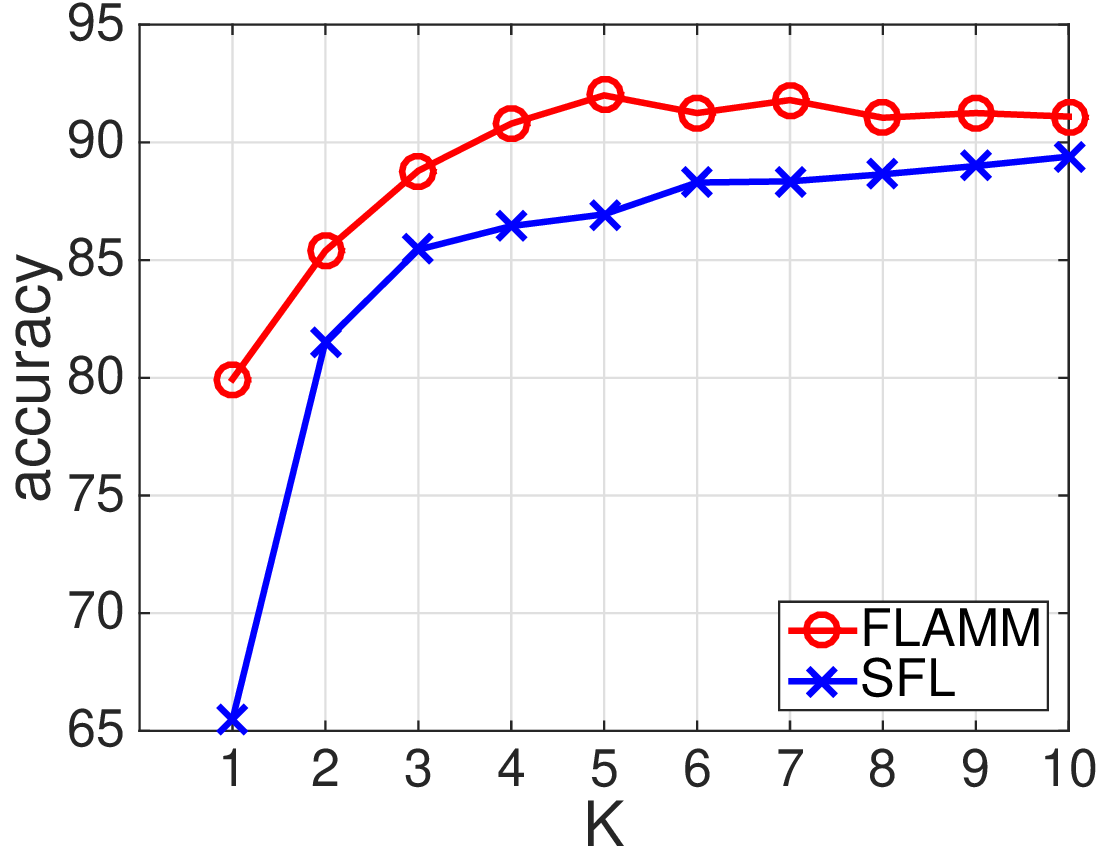}
    }
    \vspace{-4mm}
    \caption{Domain adaptation performances with different number of layers on different datasets. The parameters for the two methods are selected on validation set. }
     \label{fig:layer}
\end{figure*} 
\begin{figure*}[!t]
  \centering
    \subfigure[Amazon review dataset (B $\rightarrow$ D)]{
    \includegraphics[scale=0.25]{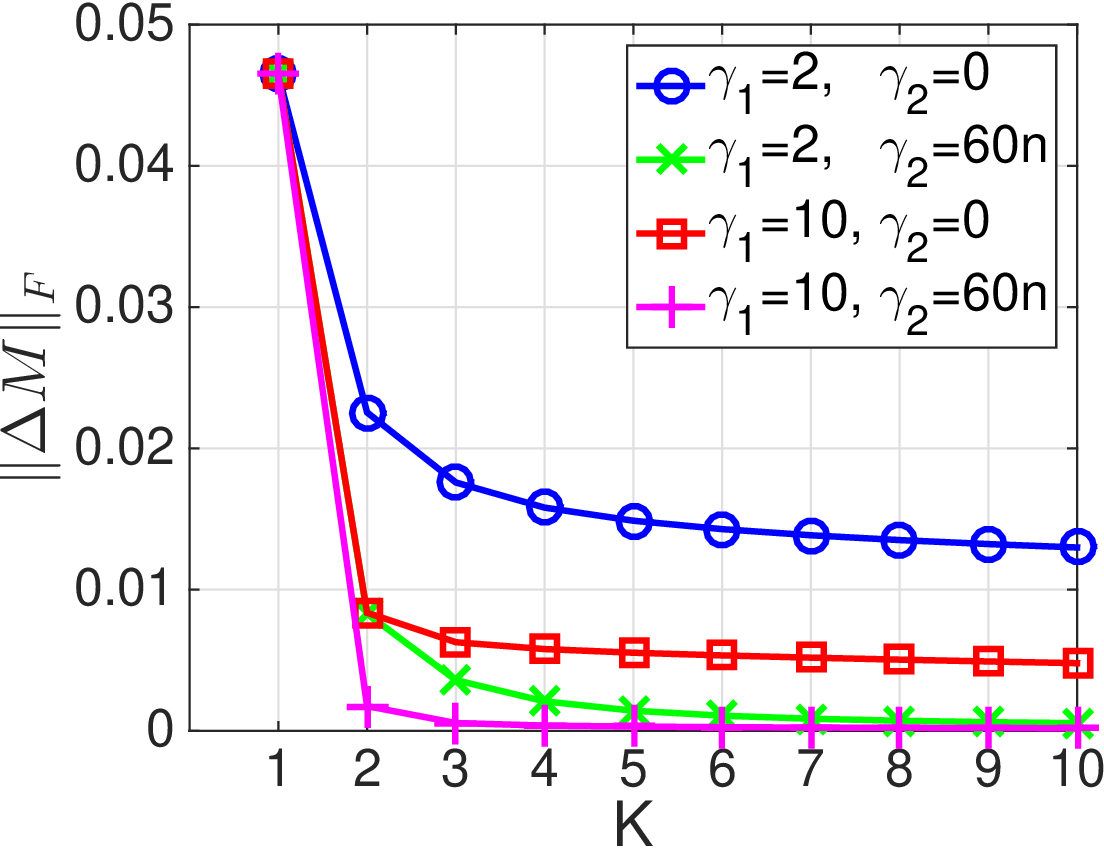}
    }\qquad\qquad
    \subfigure[Spam dataset (Public $\rightarrow$ U0)]{
    \includegraphics[scale=0.25]{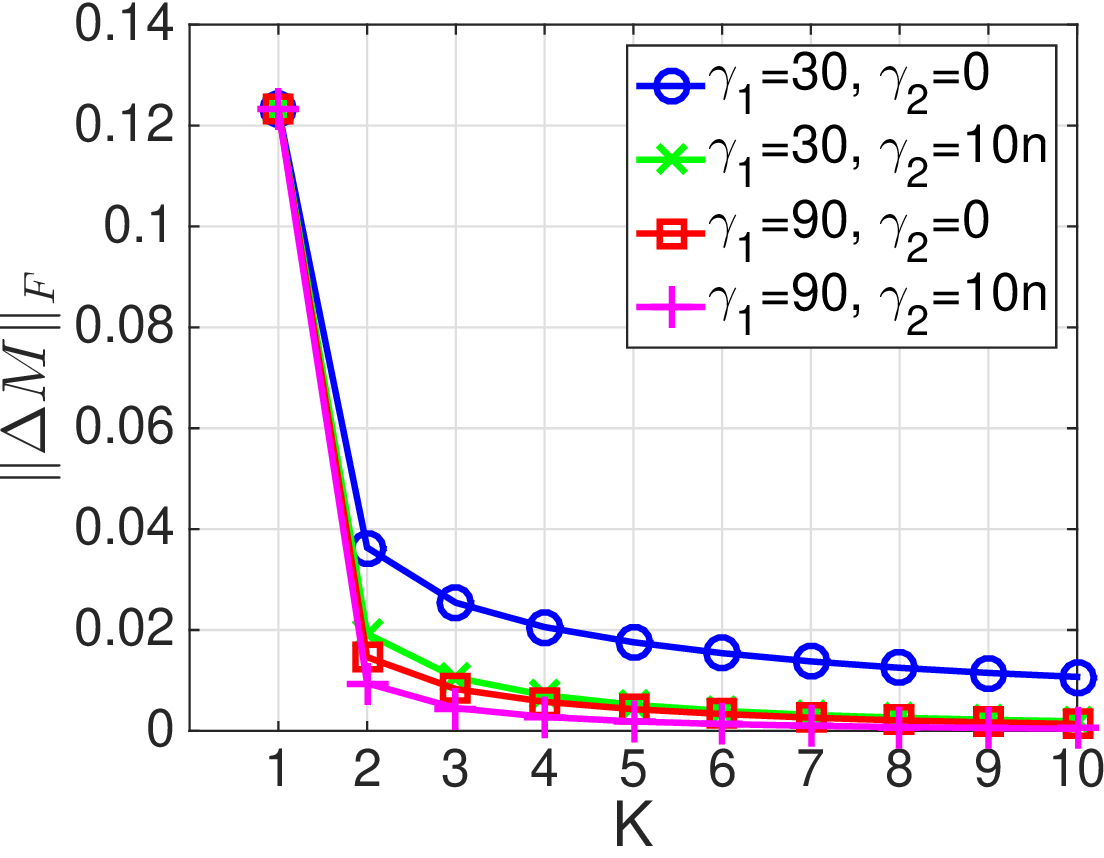}
    }
    \vspace{-4mm}  
    \caption{Distances between the second moments of source distribution and target distribution through layers on different datasets  with different $\gamma_1$ and $\gamma_2$. The distances are computed with the input data matrix of each layer. For layer 1, it is just the difference between the second moments on the original representations.}
     \label{fig:dist_layer}
     \vspace{-5mm}
\end{figure*}

\subsection{Comparison and Analysis}
For domain adaptation tasks,  traditional cross validation can not be used to select parameters since the source distribution and target distribution are different. In our experiments, we simply use a small validation set containing only 500 labeled samples selected randomly from target domain to select parameters for all feature learning algorithms~\footnote{The method we used was called STV in \cite{Zhong2010}. And it works quite well in our setting. Transfer cross validation \cite{Zhong2010} should provide similar results.}. Once we have the new  learned features, we treat the cross domain classification task as traditional supervised learning problem and the validation set was not used to select parameters for classifiers.

We report the results of two baseline representations. The first one is just the raw \textit{tf-idf} representation and the second one is the PCA representation. For PCA, the subspace is obtained from both source domain samples and target domain samples.  Besides these two baselines, we also compare our method with CORAL \cite{Sun2015} and CODA~\cite{Chen2011e}. CORAL is an effective shallow feature learning method and has been shown outperform many well-known methods, e.g. GFK \cite{Gong2012}, SVMA \cite{Duan2012} and SCL \cite{Blitzer2006}, mainly on image domain adaptation tasks. CODA is a state-of-the-art domain adaptation algorithm based on co-training. At last, we compare our method with deep learning methods. Since mSDA is better than SDA as shown in~\cite{Chen2012a}, we only provide comparisons with mSDA. We use binary representations for mSDA as in \cite{Chen2012a}, and for the other methods, samples are represented with \textit{tf-idf} and normalized to have unit length. For representations learned by all feature learning methods, we train a linear SVM on the source domain data and test it on the target domain. The performance metric is classification accuracy.

We denote our method with $K$ layers as $\textrm{FLAMM}_{K}$ and similar notations are also used for SFL and mSDA. We set $K$ as 2 for the Amazon review dataset and 5 for the spam dataset. To provide better understanding of our algorithm, we also provide results with only one layer. Parameters for representation learning process were selected based on the validation set. On the Amazon review dataset, both $\gamma_1$ and $\gamma_2$ were selected from $[10, 20, \cdots, 100]$. On the spam dataset,  they were selected from $[10, 30, 50,\cdots, 150]$.  For mSDA, the noise was selected from $[0.5, 0.6, 0.7, 0.8, 0.9]$ on the Amazon dataset and $[0.9, 0.91, \cdots, 0.99]$ on the spam dataset. For PCA, the reduced dimensionality was selected from $[50, 100 ,150,200,250, 300]$ for both datasets. For CODA\footnote{Source code: {\url {http://www.cse.wustl.edu/~mchen/code/coda.tar}}}, we used the same parameters as in \cite{Chen2011g}, except that we set $\gamma = 0.0001$ on the spam dataset. The results of CODA depend on initializations, hence we run CODA 10 times and provide the average accuracies. The results are presented in Tables~\ref{acc_on_sentiment} and \ref{acc_on_spam}. We can see that our method achieved the best performance on the two datasets. Results of FLAMM are better than those of SFL, and results with multiple layers are usually better than results with only one layer. We can see that the proposed objective and multi-layer structure did help improve the adaptation performance. 

The performances with different number of layers on the two datasets are also plotted in Figure~\ref{fig:layer}. We can see that our method is better than SFL  consistently on both datasets. On the Amazon dataset, our method become worse when  $K$ is larger than 4. The reason might be that the difference between the learned representations and the original representations is unnecessarily high.

We also present the distance between the second moments of source domain and target domain through layers with different $\gamma_1$ and $\gamma_2$ \footnote{In our implementation, $\gamma_2$ is the product of a number and $n$, where $n$ is the number of samples.} when $K=10$ as shown in Figure~\ref{fig:dist_layer}. We can see that the distance decreases through layers for both FLAMM and SFL and the distance of FLAMM is smaller than that of SFL with the same $\gamma_1$. This observation explained why SFL also improved the adaptation performance. For FLAMM, the parameters control how small domain distance we can get. Bigger $\gamma_1$ and $\gamma_2$ will lead to smaller distance between the two second moments. But at the same time, the learned representations will be more different from the original input\footnote{FLAMM tries to reconstruct the new input data matrix in each layer, which is the output data matrix from previous layer. Hence, the outputs of each layer  will  usually have larger and larger difference with the original samples through layers. }, which might not be suitable for classifiers. Hence, our method could be understood as a \textit{trade-off} between domain distance and reconstruction error.



\vspace{-3mm}
\section{Conclusions}
\vspace{-1mm}
In this paper, theoretic analysis of the factors that affect the performance of feature learning for domain adaptation was provided.  We found that the distance between the second moments of the source domain and target domain should be small, which is important for a linear classifier to generalize well on the target domain.  Based on our analysis, an extremely easy yet effective feature learning algorithm was proposed. Furthermore, our algorithm was extended by leveraging multiple layers, leading to a deep linear model. Meanwhile, we also explained why the simple ridge regression based method which does not minimize the gap between source and target distributions can improve the adaptation performance. The experimental results on the Amazon review and spam datasets corroborated the advantages of our proposed feature learning approach.
\vspace{-3mm}
\section*{Acknowledgments}
\vspace{-1mm}
This work was partially supported by the following grants: NSF-IIS 1302675, NSF-IIS 1344152, NSF-DBI 1356628 and RGC GRF Grant No. PolyU 152039/14E.
\bibliographystyle{named}
\bibliography{domain-adaptation-distance}

\end{document}